\documentclass[journal]{IEEEtran}

\ifCLASSINFOpdf
\else
   \usepackage[dvips]{graphicx}
\fi
\usepackage{url}

\usepackage{colortbl}
\usepackage{graphicx}
\usepackage{tikz}
\usepackage{amssymb}
\usetikzlibrary{arrows.meta, positioning, calc}

\usepackage[backend=biber,style=ieee,sorting=none,sortcites=true,citestyle=numeric-comp]{biblatex}

\usepackage{fancyhdr}
\usepackage{lipsum} 

\begin{document}

\title{Beyond Augmentation: Leveraging Inter-Instance Relation in Self-Supervised Representation Learning}

\author{Ali Javidani, Babak Nadjar Araabi, and Mohammad Amin Sadeghi 
\thanks{Ali Javidani, and Babak Nadjar Araabi are with the School of Electrical and Computer Engineering, College of Engineering, University of Tehran, Tehran, Iran (e-mail: \{alijavidani, araabi\}@ut.ac.ir). Mohammad Amin Sadeghi is with the Qatar Computing Research Institute, Hamad bin Khalifa University, Doha, Qatar (e-mail: msadeghi@hbku.edu.qa).(Corresponding author: Babak Nadjar Araabi)}}

\markboth{IEEE SIGNAL PROCESSING LETTERS, VOL. 32, pp. 3730-3734, 2025. doi: 10.1109/LSP.2025.3610549}
{Shell \MakeLowercase{\textit{et al.}}: Bare Demo of IEEEtran.cls for IEEE Journals}

\fancypagestyle{firstpage}{
	\fancyhf{}
	\renewcommand{\headrulewidth}{0pt}
	\fancyfoot[L]{\footnotesize
		© IEEE 2025. Personal use of this material is permitted. 
		Permission from IEEE must be obtained for all other uses, 
		in any current or future media, including reprinting/republishing 
		this material for advertising or promotional purposes, 
		creating new collective works, for resale or redistribution to servers or lists, 
		or reuse of any copyrighted component of this work in other works.}
}

\maketitle
\thispagestyle{firstpage}
\begin{abstract}
	This paper introduces a novel approach that integrates graph theory into self-supervised representation learning. Traditional methods focus on intra-instance variations generated by applying augmentations. However, they often overlook important inter-instance relationships. While our method retains the intra-instance property, it further captures inter-instance relationships by constructing $k$-nearest neighbor (KNN) graphs for both teacher and student streams during pretraining. In these graphs, nodes represent samples along with their latent representations. Edges encode the similarity between instances. Following pretraining, a representation refinement phase is performed. In this phase, Graph Neural Networks (GNNs) propagate messages not only among immediate neighbors but also across multiple hops, thereby enabling broader contextual integration. Experimental results on CIFAR-10, ImageNet-100, and ImageNet-1K demonstrate accuracy improvements of 7.3\%, 3.2\%, and 1.0\%, respectively, over state-of-the-art methods. These results highlight the effectiveness of the proposed graph-based mechanism. The code is publicly available at \url{https://github.com/alijavidani/SSL-GraphNNCLR}.
\end{abstract}


\IEEEpeerreviewmaketitle

\section{Introduction}

\IEEEPARstart{S}{elf-supervised} representation learning has become a powerful means of extracting features from unlabeled data, bridging the gap between unsupervised and supervised paradigms. By employing pretext tasks \cite{RN267, RN326, RN218, RN343}, these methods exploit data structure to lessen reliance on labeled datasets. Contrastive approaches such as SimCLR \cite{RN46} and MoCo \cite{RN44} focus on negative samples, while non-contrastive ones, including BYOL \cite{RN38}, SimSiam \cite{RN47}, and DINO \cite{RN43}, align augmented views via self-distillation without explicit negatives. Extensions explore patch-level processing \cite{RN357, RN331, RN226, RN348, RN171}, clustering strategies \cite{RN25,RN45,RN99}, and information maximization \cite{RN252,RN260,RN269,RN310,RN328,RN79} to capture global/local features and promote diversity. Consequently, self-supervised methods now rival supervised ones in many tasks, showing strong versatility and transferability \cite{RN342, RN344, RN345, RN358}. Broader surveys further review transfer and self-supervised learning, outlining definitions, applications, and limitations \cite{RN359}. Related studies have also explored adversarial and structured self-supervised formulations to improve robustness in tasks such as remote sensing object detection~\cite{RN360}.

The baseline SimCLR method applies contrastive learning by generating two augmented views of a sampled image, pulling their representations closer while pushing them away from others~\cite{RN46}. However, common augmentations (e.g., color jittering, cropping, Gaussian blur) mainly capture intra-sample variations while preserving overall structure, and thus fail to introduce the broader inter-sample diversity—such as differences in background, lighting, or pose—needed for robust feature learning. The NNCLR framework~\cite{RN214} addresses this by selecting the nearest neighbor of an image as its positive pair. Such a neighbor typically belongs to the same semantic class but appears under distinct, naturally occurring conditions (e.g., the same dog breed photographed indoors versus outdoors). Aligning these representations enables the network to learn features that generalize across real variations rather than relying solely on artificial transformations.

NNCLR outperforms traditional contrastive methods but has two main shortcomings~\cite{RN214}. First, it selects nearest neighbors from latent representations that evolve during pretraining, often resulting in imprecise choices during early epochs when the network is immature. Second, it discards augmentation-based supervisory signals, despite evidence of their benefit to self-supervised learning~\cite{RN309, RN259}. AdaSim addresses these issues by adaptively switching between two modes~\cite{RN237}: relying on augmentation signals in the initial epochs when neighbors are unreliable, and later sampling positives from the stabilized nearest-neighbor distribution.

Although AdaSim leverages a distribution of nearest neighbors from recent epochs, it samples only a single neighbor for each anchor~\cite{RN237}. This overlooks other relevant neighbors that could provide varied contexts and refine a sample’s representation. Unlike methods focusing on intra-sample relationships—through input augmentations~\cite{RN43} or informative patches \emph{within} each image~\cite{RN357, RN113}—our approach emphasizes inter-sample relationships by refining representations \emph{across} images through information propagation. Specifically, we construct a $k$-nearest neighbor (KNN) graph that encodes the most similar samples via directed edges, dynamically updated throughout pretraining to capture evolving neighbor relationships. By aggregating information from multiple neighbors, each sample’s representation is refined more effectively, benefiting from broader semantic diversity.

This graph-based perspective offers several advantages:  
(1) it provides a unified structure to capture inter-sample relationships, allowing each sample to leverage information from multiple neighbors;  
(2) by modeling higher-order connectivity, it propagates semantic information across multiple hops, improving generalization for visually diverse samples within the same class;  
(3) it supports a wide spectrum of graph-theoretic and deep learning approaches—from classical metrics (e.g., centrality and clustering) to graph neural networks (GNNs)—to fuse relational context and yield richer embeddings.

\section{Proposed Method}
\label{sec:proposed_method}
\subsection{Proposed Overall Framework}
Our framework consists of two phases: representation extraction (Fig.~1) and representation refinement (Fig.~3). In the extraction phase, an image undergoes two augmentations and is processed by student and teacher networks in a self-distillation setup to obtain backbone representations (Section~\ref{backbone_representation}). Based on inter-sample similarities, separate KNN graphs are then constructed for the teacher and student streams (Section~\ref{knn_construction}). These representations are projected into a higher-dimensional space via an MLP projector, and a cross-entropy loss aligns them (Section~\ref{loss_function_calculation}). At the end of this phase, the KNN graphs preserve node attributes as sample embeddings and edges as inter-sample relationships. In the refinement phase, the KNN graphs are further processed to improve embeddings through a self-supervised node representation learning procedure, where GNN layers enable message passing among connected nodes. Consequently, the representations are refined via multi-hop interactions, producing robust and semantically rich embeddings (Section~\ref{representation_refinement}).

\begin{figure}[t]
	\centerline{\includegraphics[width=\columnwidth]{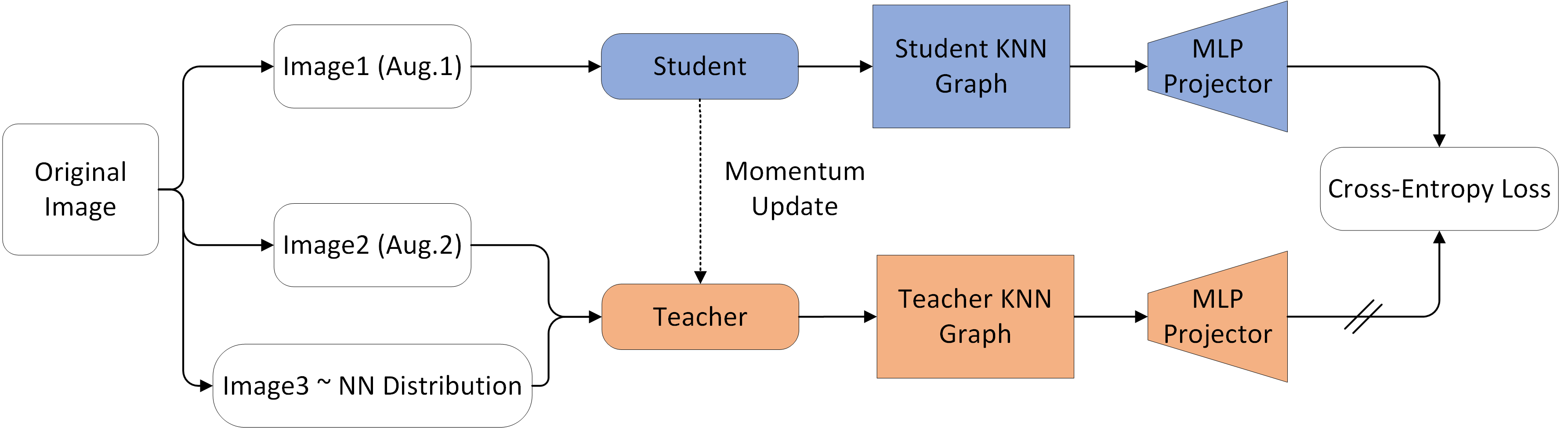}}
	\caption{Illustration of the proposed self-distillation framework with KNN graphs. The original image is augmented to produce two views. The student encoder processes one view, and the teacher encoder processes the other. Each stream constructs a KNN graph based on the latent representations. Finally, a cross-entropy loss aligns the teacher and student embeddings after they are projected to a higher-dimensional space via MLP projectors.}
	
	\label{fig1}
\end{figure}

\subsection{Backbone Representations}
\label{backbone_representation}
Similar to DINO \cite{RN43} and AdaSim \cite{RN237}, we adopt a teacher-student self-distillation framework. In DINO \cite{RN43}, a single image is augmented twice: one augmentation is fed to the student, and the other to the teacher network. Building on this idea, in AdaSim an adaptive mechanism is introduced to determine whether both networks receive augmented images or if the teacher network should instead process a sample’s nearest neighbor \cite{RN237}. Specifically, if the network is not yet sufficiently mature, it reverts to the DINO-style approach of using two augmented images (Image 2 in Fig.~\ref{fig1}); otherwise, a sample from the nearest neighbor distribution is drawn and provided to the teacher (Image 3 in Fig.~\ref{fig1}). Network maturity is assessed by checking whether the nearest neighbor is the same as the original sample; if so, the model is deemed adequately trained. 

For each sample \(v_i\), we compute a similarity distribution over candidate neighbors. Let 
$\mathcal{N}_e^{(\tau)}(v_i)$ denote the set of top \(e\) nearest neighbors of \(v_i\) in epoch \(\tau\), as determined by cosine similarity. We aggregate the similarity scores across the \(w\) most recent epochs as
\begin{equation}
	S(v_i, v_j) = \sum_{\tau=1}^{w} \mathbf{1}\{v_j \in \mathcal{N}_e^{(\tau)}(v_i)\} \, \cos(\theta_{ij}),
	\label{eq:sim_scores}
\end{equation}
where \(\mathbf{1}\{\cdot\}\) is the indicator function and \(\cos(\theta_{ij})\) denotes the cosine similarity between \(v_i\) and \(v_j\). The similarity distribution is then given by the softmax function:
\begin{equation}
	p(v_j \mid v_i) = \frac{\exp\bigl(S(v_i,v_j)\bigr)}{\sum_{v_k \in \mathcal{N}(v_i)} \exp\bigl(S(v_i,v_k)\bigr)},
	\label{eq:sim_dist}
\end{equation}
with
\begin{equation}
\mathcal{N}(v_i) = \bigcup_{\tau=1}^{w} \mathcal{N}_e^{(\tau)}(v_i).
\end{equation}
This procedure aggregates similarity information across \(w\) epochs and normalizes the scores to form a probability distribution over the candidate neighbors.

\subsection{Proposed K-Nearest Neighbor Graph Construction}
\label{knn_construction}
Following the processing of images through the student and teacher networks, and prior to projecting the backbone representations into a higher-dimensional space via the MLP projector, we construct the graph structures (see Fig.~\ref{fig1}). Specifically, we build a directed $k$-nearest neighbor (KNN) graph $G = (V,E)$ for each stream. Here, each node represents a distinct sample from the dataset, encompassing both training and validation sets, such that \(|V| = N\), where \(N\) denotes the total number of samples. An edge \((v_i, v_j) \in E\) is established if \(v_j\) is among the top \(k\) nearest neighbors of \(v_i\) in the precomputed similarity distribution, under the assumption that \(k < e\). Formally, 
\begin{equation}
(v_i, v_j) \in E \quad iff \quad v_j \in \mathrm{top}_k\bigl(\mathcal{N}(v_i)\bigr)
\end{equation}
where $\mathcal{N}(v_i)$ denotes the set of candidate neighbors for $v_i$ and $\mathrm{top}_k(\cdot)$ returns the $k$ most similar samples. To avoid self-loops, if $v_i$ is found among its own top-$k$ neighbors, the next closest node is selected instead. By repeating this procedure over the most recent $w$ epochs, each node accumulates $k \times w$ directed edges, effectively capturing the evolving inter-sample relationships in the latent space. This step enables the model to capture neighbor relationships in the lower-dimensional latent space. It preserves crucial inter-sample information without disrupting the core self-distillation process.

To store and update nearest neighbor relationships over multiple epochs, we adopt a circular queue mechanism in a compact $2 \times (N \times k \times w)$ array, where the first row holds source nodes and the second row holds their corresponding $k$ nearest neighbors for each epoch. The array maintains node indices across $N$ samples, $k$ neighbors, and $w$ recent epochs. When new edges arrive for the current epoch, they overwrite the oldest entries, preserving only the most recent relationships. An example layout is shown in Table~\ref{table0} for $k = 2$ and $w = 3$, where each set of two columns corresponds to a single epoch’s edges, and blocks of columns are overwritten in a rolling fashion (e.g., edges from epoch $(\tau + w) \bmod w$ replace those from epoch $\tau \bmod w$).

\begin{table}[t!]
	\caption{Visualization of the circular queue mechanism with a $2 \times (k \times w)$ array for $k=2$ and $w=3$. $d_{1}^{(\tau)}$ \& $d_{2}^{(\tau)}$ denote the two nearest neighbors of source node "src" for a given epoch $\tau$.}
	\label{table0}
	\centering
	\begin{tabular}{c|cc|cc|cc}
		\hline
		& \multicolumn{2}{c|}{\scriptsize $\tau \bmod w=0$}
		& \multicolumn{2}{c|}{\scriptsize $\tau \bmod w=1$} 
		& \multicolumn{2}{c}{\scriptsize $\tau \bmod w=2$} \\ \hline
		\scriptsize Source & \scriptsize src & \scriptsize src 
		& \scriptsize src & \scriptsize src 
		& \scriptsize src & \scriptsize src \\ \hline
		\scriptsize Dest. & 
		\scriptsize $d_{1}^{(\tau)}$ & \scriptsize $d_{2}^{(\tau)}$ 
		& \scriptsize $d_{1}^{(\tau+1)}$ & \scriptsize $d_{2}^{(\tau+1)}$ 
		& \scriptsize $d_{1}^{(\tau+2)}$ & \scriptsize $d_{2}^{(\tau+2)}$ \\ \hline
	\end{tabular}
\end{table}

\subsection{MLP Projection and Loss Function Calculation}
\label{loss_function_calculation}

After graph construction, the encoded features from both teacher and student are projected to a higher-dimensional space via a 3-layer MLP. To stabilize training, the teacher output is centered and sharpened, while the student output is only sharpened. Centering removes the batch mean, and sharpening applies temperature scaling to reduce entropy. A cross-entropy loss is then used to align the two distributions:
\begin{equation}
	\ell_i = -\, \mathbf{t}_i^{\top} \log \mathbf{s}_i,
	\label{eq:ce_loss}
\end{equation}
where $\mathbf{t}_i$ and $\mathbf{s}_i$ denote the teacher and student output vectors for sample $i$.

\subsection{Representation Refinement with Graph Neural Networks}
\label{representation_refinement}
In NNCLR~\cite{RN214} and AdaSim~\cite{RN237}, only the nearest neighbor—or a sample drawn from its distribution—contributes to a sample’s representation. By contrast, our method constructs $k$-nearest neighbor (KNN) graphs to systematically integrate information from multiple neighbors, ensuring that each anchor benefits from richer context. Applying graph neural network (GNN) layers further enhances this process: through message passing, nodes aggregate not only direct but also higher-order neighborhood information, which mitigates representation collapse, strengthens semantic consistency across local neighborhoods, and encourages smoother manifolds in the embedding space. These effects are theoretically grounded in the view that GNN message passing performs local regularization on the data manifold, thereby improving robustness and discriminability in the self-supervised setting.

In each GNN layer, a node’s embedding is updated by integrating information from its neighbors. For a node \(v\) at layer \(l+1\) with neighbor set \(\mathcal{N}(v)\) and previous embedding \(h_v^{(l)}\), an aggregation function (e.g., sum, mean, or max) first collects neighbor embeddings:
\begin{equation}
	m_v^{(l+1)} = \mathrm{AGG}\bigl(\{h_u^{(l)} : u \in \mathcal{N}(v)\}\bigr),
	\label{eq:gnn_agg}
\end{equation}
producing a message \(m_v^{(l+1)}\). An update function then combines \(h_v^{(l)}\) with \(m_v^{(l+1)}\) to yield the new embedding:
\begin{equation}
	h_v^{(l+1)} = \mathrm{UPDATE}\bigl(h_v^{(l)},\, m_v^{(l+1)}\bigr).
	\label{eq:gnn_update}
\end{equation}
In practice, the update is typically a linear or MLP transformation with a non-linear activation. Regardless of the choice of \(\mathrm{AGG}\) and \(\mathrm{UPDATE}\), the principle remains the same: neighbors pass messages to each node, which updates its embedding by integrating those messages with its current state.

To further improve representation quality, we employ Jumping Knowledge (JK), which aggregates outputs from multiple layers instead of relying only on the last layer~\cite{RN350}. Using strategies such as concatenation, summation, or maximization, JK captures multi-scale neighborhood information, alleviates over-smoothing in deeper GNNs, and enables richer feature extraction across receptive fields.

Both teacher and student graphs are constructed from the same dataset but differ in how they capture sample relationships, leading their node representations to converge during training. This setting is consistent with self-supervised graph node representation learning, where methods such as SelfGNN~\cite{RN340} and BGRL~\cite{RN341} learn embeddings without labels. In these approaches, a graph is typically augmented (e.g., by edge dropping or feature perturbation), and separate GNNs process each view, with an alignment loss synchronizing the resulting embeddings.

\begin{figure}[t]
	\centerline{\includegraphics[width=\columnwidth]{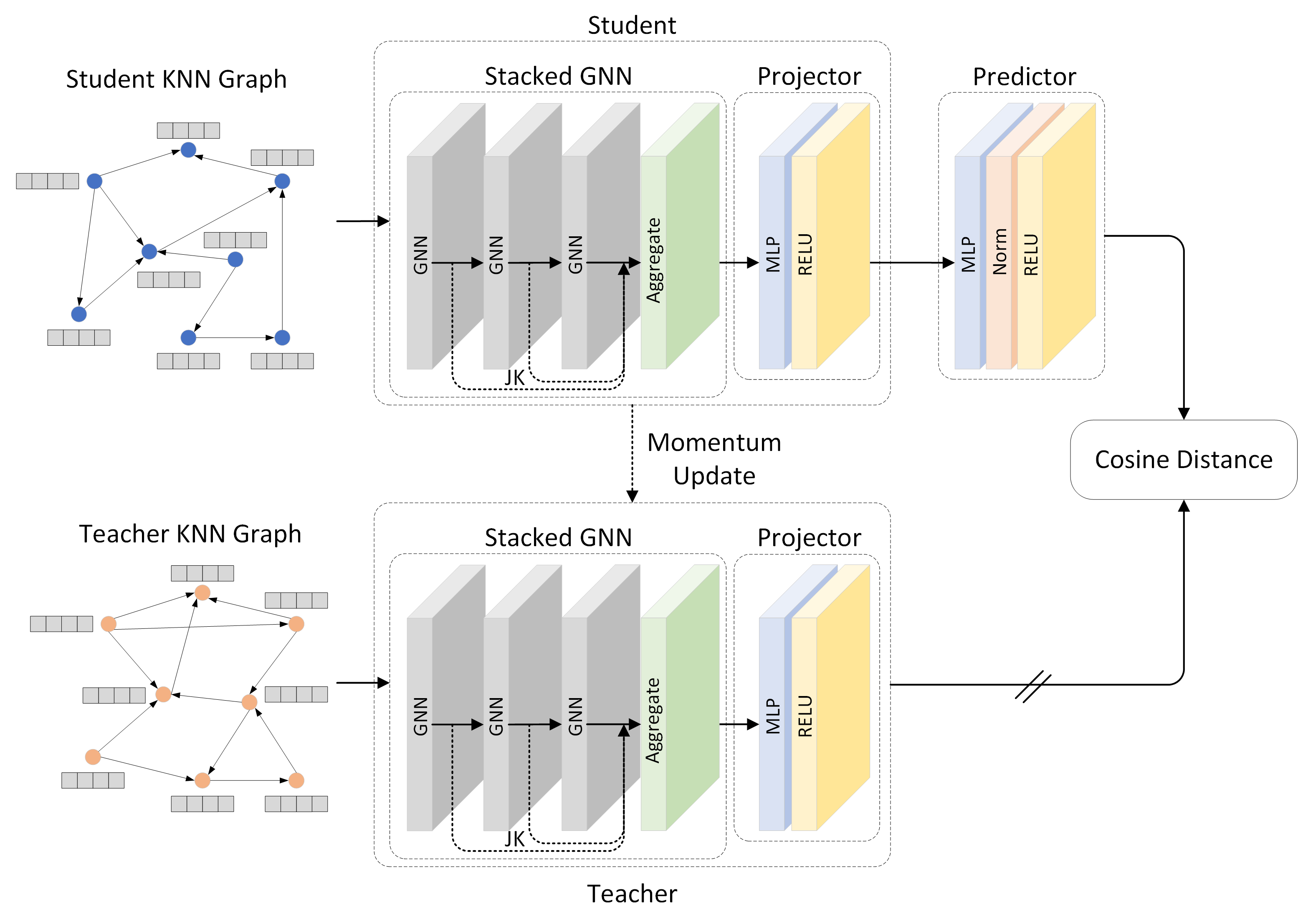}}
	\caption{Overview of the representation refinement phase. Teacher and student graphs are processed by separate GNNs, with teacher parameters updated via an exponential moving average of the student. A cosine distance loss encourages consistency between embeddings. Dashed arrows represent Jumping Knowledge (JK), which aggregates outputs from different GNN layers based on the selected strategy (e.g., concat, sum, or max).}
	\label{fig3}
\end{figure}

Our framework (Fig.~\ref{fig3}) removes the need for artificial graph augmentation by naturally providing two complementary views: teacher and student KNN graphs. Separate GNNs refine node embeddings in each view while preserving the structural relationships, and alignment is enforced by minimizing cosine distance. Specifically, for a node $i$ with student embedding $\mathbf{s}_i$ and teacher embedding $\mathbf{t}_i$, the distance is

\begin{equation}
	\ell_i = 2 - 2\,\frac{\mathbf{s}_i \cdot \mathbf{t}_i}{\|\mathbf{s}_i\| \,\|\mathbf{t}_i\|}.
	\label{eq:cosine_loss}
\end{equation}

The teacher network is updated through an exponential moving average of the student parameters, preserving the fully self-supervised setting. Training further refines node embeddings via multi-hop message passing across three GNN layers (Fig.~\ref{fig3}), allowing each sample to incorporate information from its immediate, second-, and third-hop neighbors. This process captures complex inter-sample dependencies and uncovers intrinsic clustering structures in the learned feature space.

\section{Experimental Results \& Discussion}
In this section, the experimental evaluation of our proposed framework is presented. We describe the training datasets, report the evaluation results, and discuss key parameters that influence the performance of our method. 

\subsection{Datasets}
We conduct experiments on three widely used image datasets: CIFAR-10 \cite{RN233}, ImageNet-100 \cite{RN234}, and ImageNet-1K \cite{RN234}. CIFAR-10 comprises 60,000 images spanning 10 classes, with 50,000 training and 10,000 validation samples. ImageNet-1K contains 1,281,167 training images, 50,000 validation images across 1,000 classes. ImageNet-100 is a balanced subset of ImageNet-1K featuring 100 randomly selected classes, each holding 1,300 training and 50 validation images.

\subsection{Implementation Details}
The backbone for representation extraction is a ViT-Base model producing 768-dimensional embeddings. All other hyperparameters follow the DINO~\cite{RN43} and AdaSim~\cite{RN237} GitHub implementations. For representation refinement, we use a learning rate of 0.0001 and momentum of 0.99. The GNN layers, configured with 768 units, are trained for 1000 epochs.

\subsection{Evaluation: Linear Image Classification}
We pre-trained our method on CIFAR-10, ImageNet-100, and ImageNet-1K in a fully self-supervised manner, without using label information. To evaluate the learned representations, we trained a logistic regression classifier on the enhanced features extracted from both the training and validation sets. Table~\ref{table1} reports the classification accuracies. The results show that our method outperforms comparable approaches. Notably, evaluation was performed solely on the refined backbone features, with the MLP projector omitted to isolate the intrinsic quality of the enhancements.

\begin{table}
	\caption{Linear evaluation accuracies percentage of the proposed and related methods using a logistic regression classifier}
	\label{table1}
	\centering
	\begin{tabular}{|c|c|c|c|}
		\hline
		\rowcolor{gray!20} 
		Method & CIFAR-10 & ImageNet-100 & ImageNet-1K \\ 
		\hline
		SimCLR \cite{RN46} & 76.1 & 75.4 & 68.9 \\
		NNCLR \cite{RN214} & 78.5 & 77.3 & 70.9 \\
		DINO \cite{RN43} & 78.6 & 77.2 & 71.3 \\
		AdaSim \cite{RN237} & 81.1 & 78.4 & 71.8 \\
		EsViT \cite{RN113} & 81.2 & 78.7 & 72.5 \\
		MSBReg \cite{RN220} & 82.4 & 80.6 & 70.5 \\ \hline
		Ours & \textbf{89.7} & \textbf{83.8} & \textbf{72.8} \\ \hline
	\end{tabular}
\end{table}

\subsection{Effect of Edges Per Node}
The number of edges per node is given by \(k \times w\), where \(k\) is the number of nearest neighbors per sample and \(w\) is the number of epochs over which these neighbors are tracked. Sparse graphs (small \(k\) and \(w\)) may fail to capture sufficient inter-sample information, while overly dense graphs (large \(k\) and \(w\)) risk introducing irrelevant edges by connecting dissimilar samples. Reducing \(k\) yields more selective edge formation that better preserves class-specific relationships. Thus, a more effective strategy is to use a smaller \(k\) and a larger \(w\). Our experiments (Table~\ref{table2}) confirm this: the best results occur with low \(k\) and high \(w\), whereas increasing \(k\) at fixed \(w\) degrades accuracy.

\begin{table}[h!]
	\caption{Effect of edges per node parameter (\(k\) \& \(w\)) on model accuracy (percentage) for CIFAR-10 and ImageNet-100}
	\label{table2}
	\centering
	\begin{tabular}{|c|c|c|c|c|}
		\hline
		\rowcolor{gray!20} 
		\( k \) & 1 & 2 & 3 & 4 \\
		\hline
		\rowcolor{gray!20} 
		\( w \) & 15 & 8 & 5 & 5 \\
		\hline
		\rowcolor{gray!20}
		Edges Per Node & 15 & 16 & 15 & 20\\
		\hline
		CIFAR-10 & \textbf{89.7} & 89.2 & 88.4 & 87.5\\		\hline
		ImageNet-100 & \textbf{83.8} & 82.9 & 80.3 & 78.4\\
		\hline
	\end{tabular}
\end{table}

\begin{table}[h!]
	\caption{Effect of GNN layer types on model accuracy (percentage) for CIFAR-10 and ImageNet-100 with $k=1$ \& $w=15$}
	\label{table3}
	\centering
	\begin{tabular}{|c|c|c|c|c|}
		\hline
		\rowcolor{gray!20} 
		Dataset & GCN & GAT & GraphSAGE & GIN \\
		\hline
		CIFAR-10  & \textbf{89.7} & 86.5 & 71.6 & 62.4\\
		\hline
		ImageNet-100  & \textbf{83.8} & 83.1 & 81.1 & 68.9\\
		\hline
	\end{tabular}
\end{table}

\begin{table}[h!]
	\caption{Effect of Jumping Knowledge (JK) and Aggregator Type on model accuracy (percentage) for CIFAR-10 and ImageNet-100}
	\label{table4}
	\centering
	\begin{tabular}{|c|c|c|c|c|}
		\hline
		\rowcolor{gray!20} 
		Dataset & JK Disabled & JK(Sum) & JK(Max) & JK(Concat) \\
		\hline
		CIFAR-10  & 85.9 & 85.0 & 84.5 & \textbf{89.7}\\
		\hline
		ImageNet-100  & 82.6 & 83.2 & 83.1 & \textbf{83.8}\\
		\hline
	\end{tabular}
\end{table}

\subsection{Effect of GNN Architecture}
We evaluate four common GNN architectures: Graph Convolutional Networks (GCN)~\cite{RN336}, Graph Attention Networks (GAT)~\cite{RN339}, GraphSAGE~\cite{RN337}, and Graph Isomorphism Networks (GIN)~\cite{RN338}. Results on CIFAR-10 and ImageNet-100 (Table~\ref{table3}) indicate that GCN and GAT outperform GraphSAGE and GIN. Their advantage likely stems from simpler and more stable aggregation—mean aggregation in GCN and attention-weighted mean in GAT—which yield robust embeddings under self-supervised, noisy graph constructions. In contrast, GraphSAGE’s complex aggregation functions and GIN’s sum aggregation tend to amplify noise, resulting in weaker performance.

\subsection{Effect of Jumping Knowledge}
We evaluated the impact of applying Jumping Knowledge (JK) in the representation refinement phase. The results presented in Table \ref{table4} show that using JK consistently improves performance compared to disabling it. Among different aggregation strategies, concatenation outperformed both mean and max, indicating its superior ability to preserve multi-layer information and enhance representation quality.

\section{Conclusion}
In this paper, we investigated the benefits of incorporating $k$-nearest neighbor graphs into self-supervised representation learning frameworks. Our approach uniquely emphasizes the capture of inter-instance relationships by refining sample embeddings through graph neural networks. These GNN layers propagate messages over multiple hops, thereby generating representations with broader contextual awareness. Experimental evaluations on CIFAR-10, ImageNet-100, and ImageNet-1K demonstrated that our method outperformed existing approaches, underscoring the effectiveness of integrating graph-based techniques into self-supervised learning.

\printbibliography

\end{document}